\documentclass[runningheads]{llncs}

\usepackage{eccv}

\usepackage{eccvabbrv}

\usepackage{graphicx}
\usepackage{booktabs}

\usepackage[accsupp]{axessibility}  

\usepackage{enumitem}

\usepackage{hyperref}

\usepackage{orcidlink}
\usepackage{bbm}

\usepackage{array}       
\usepackage{multirow}    
\usepackage{colortbl}    
\usepackage{xcolor}      
\usepackage{makecell}    

\usepackage{wrapfig}

\begin{document}

\title{TIR-Agent: Training an Explorative and Efficient Agent for Image Restoration} 

\titlerunning{TIR-Agent}

\author{Guoli Jia\inst{1}$^\star$ \and
Yisheng Zhang\inst{1}$^\star$ \and
Haote Hu\inst{1}$^\star$ \and
Shanxu Zhao\inst{1} \and
Kaikai Zhao\inst{1,2} \and
Long Sun\inst{3} \and
Xinwei Long\inst{1} \and
Kai Tian\inst{1,4} \and
Che Jiang\inst{1,4} \and
Zhaoxiang Liu\inst{2} \and
Kai Wang\inst{2} \and
Shiguo Lian\inst{2} \and
Kaiyan Zhang\inst{1,4}$^\dagger$ \and
Bowen Zhou\inst{1,5}$^\dagger$
}

\authorrunning{Jia et al.}

\institute{Tsinghua University, Beijing, China \and
Data Science \& Artificial Intelligence Research Institute, China Unicom, Beijing, China \and
Hunan University, Changsha, China \and
Frontis.AI, Beijing, China \and
Shanghai Artificial Intelligence Laboratory, Shanghai, China\\
}

\makeatletter
\def\blfootnote#1{%
  \begingroup
  \renewcommand\thefootnote{}%
  \renewcommand\@makefnmark{}%
  \footnotetext{#1}%
  \addtocounter{footnote}{-1}%
  \endgroup
}
\makeatother

\maketitle

\blfootnote{$\star$: Equal contribution. Guoli Jia (exped1230@gmail.com) leads the project. $\dagger$: Corresponding authors.}

\begin{abstract}
  Vision-language agents that orchestrate specialized tools for image restoration (IR) have emerged as a promising method, yet most existing frameworks operate in a training-free manner.
  They rely on heuristic task scheduling and exhaustive tool traversal, resulting in sub-optimal restoration paths and prohibitive computational cost. 
  We argue that the core bottleneck lies in the absence of a learned policy to make decision, as a vision-language model cannot efficiently handle degradation-aware task ordering and tool composition. 
  To this end, we propose TIR-Agent, a trainable image restoration agent that performs a direct tool-calling policy through a two-stage training pipeline of supervised fine-tuning (SFT) followed by reinforcement learning (RL).
  Two key designs underpin effective RL training: (i) a random perturbation strategy applied to the SFT data, which broadens the policy's exploration over task schedules and tool compositions, 
  and (ii) a multi-dimensional adaptive reward mechanism that dynamically re-weights heterogeneous image quality metrics to mitigate reward hacking.
  To support high-throughput, asynchronous GPU-based tool invocation during training, we further develop a globally shared model-call pool.
  Experiments on both in-domain and out-of-domain degradations show that TIR-Agent outperforms 12 baselines, including 6 all-in-one models, 3 training-free agents, and 3 proprietary models, and achieves over 2.5$\times$ inference speedup by eliminating redundant tool executions.
\end{abstract}

\section{Introduction}
\label{sec:intro}
Images captured in real-world scenarios inevitably suffer from composite degradations, \textit{i.e.}, mixtures of noise, low light, compression artifacts, and more~\cite{aio_airnet, iragent_depictqa}. 
This compositional nature makes the image restoration (IR) a challenging sequential decision-making problem that goes well beyond the scope of previous single degradation restoration network~\cite{ir_swinir, ir_nafnet}.

Two dominant paradigms have been developed to address composite degradations.
All-in-one models~\cite{jiang2024autodir, aio_airnet, aio_dc} employ a single network to jointly handle multiple degradations within one forward pass.
While conceptually simple, sharing parameters across diverse degradations forces the model to trade off capacity, resulting in degraded performance on under-represented or long-tail degradations~\cite{guo2024parameter, iragent_greedy}.
Agent-based frameworks \cite{iragent_depictqa, zuo2025kagent} take a different approach.
They treat pre-trained specialist restoration models as independent tools and leverage a vision-language model (VLM) \cite{iragent_depictqa, depictqa_v2} to dynamically select and compose them. 
This modular design decouples capability from capacity, offering superior scalability and adaptability to unseen degradation combinations.

\begin{figure}[!t]
	\centering
	\includegraphics[width=1.0\textwidth]{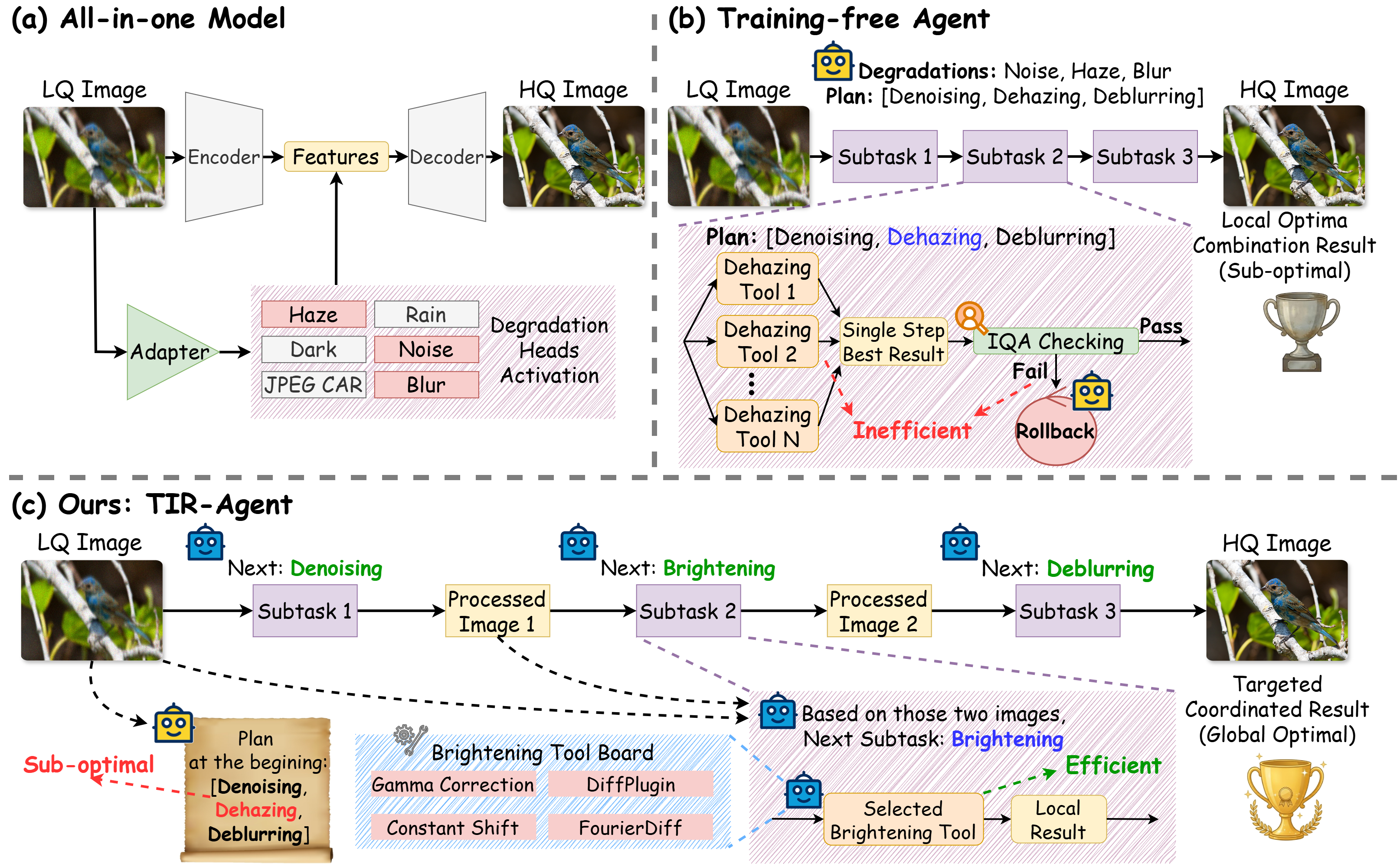}
	\caption{\textbf{Comparison of restoration paradigms.} \textbf{(a) All-in-one model:} Uses a single network with specialized heads to inject various degradation features. \textbf{(b) Training-free Agent:} Uses VLMs for static task planning, executing sub-tasks sequentially via exhaustive tool traversal. \textbf{(c) TIR-Agent (Ours)}: A trainable agent that dynamically decides the next sub-task based on current results and learns to directly select the optimal tool, avoiding redundant execution.}
	\label{intro}
\end{figure}

Despite their promise, current agent-based frameworks almost exclusively operate in a training-free setting \cite{iragent_depictqa, iragent_restoreagent, iragent_chain, jiang2025multi, iragent_hybrid, zuo2025kagent}, which intrinsically faces three limitations:
(i) \textbf{Sub-optimal task scheduling}. 
Without learning from experience, training-free agents rely on large language model (LLM)-based reasoning or fixed rules to determine the execution order of IR tasks~\cite{iragent_chain, jiang2025multi, iragent_hybrid}. 
Such static strategies fail to capture the complex, input-dependent interactions between degradations.
For example, the optimal ordering of denoising and dehazing may vary substantially across images.
(ii) \textbf{Inefficient tool selection}. 
Given an IR task (\textit{e.g.}, denoising), training-free agents typically exhaustively execute all candidate tools, such as X-Restormer~\cite{chen2024comparative}, Restormer~\cite{zamir2022restormer}, and NAFNet~\cite{ir_nafnet}, then selecting the best result~\cite{iragent_depictqa, zuo2025kagent}.
This enumeration scales linearly with the tool size, making inference prohibitively expensive.
(iii) \textbf{Absence of reference-guided optimization}. 
Operating purely at inference time, training-free agents can only rely on no-reference (NR) image quality metrics~\cite{simplecall, iragent_restoreagent} to compare results. 
Without access to full-reference (FR) supervision during the learning phase, the fidelity of restored images is inherently limited.
Recent advances in agentic reinforcement learning (RL) have demonstrated that training LLM-based agents to interact with tools, rather than relying on static prompting, leads to significantly more robust and efficient policies~\cite{chan2024mlebench, ferrag2025llm}.
A key insight from this emerging paradigm is that the agent's tool-use strategy should itself be a learned behavior, optimized through trial-and-error interaction based on the feedback.
IR has well-defined reward signals that can be computed from FR and NR image quality metrics for training. 
Yet, to our knowledge, no prior work has systematically applied agentic RL to train a vision-language agent for IR.
We present TIR-Agent, a trainable image restoration agent that learns to directly invoke the suitable tool at each step, eliminating the need for exhaustive search. 
TIR-Agent is trained through a two-stage pipeline. 
In the supervised fine-tuning (SFT) stage, the VLM learns a base policy from demonstration data.
To prevent the policy from converging to narrow, sub-optimal patterns, we apply a exploration-driven perturbation (EDP) strategy that injects controlled perturbation into the SFT trajectories, encouraging the model to explore a broader space of task schedules and tool compositions. 
In the subsequent RL stage, the agent refines its policy by interacting with the FR and NR feedback.
We observe that naively combining multiple image quality metrics as a reward signal leads to reward hacking, \textit{i.e.}, the agent exploits partial metrics at the expense of others.
To address this issue, we propose a multi-dimensional adaptive reward (MAR) mechanism that dynamically adjusts the weight of each metric based on training progress, guiding the agent toward globally balanced optimization.
To enable efficient RL training with GPU-intensive visual models, we further develop a distributed, asynchronous infrastructure centered on a globally shared model-call pool (MC-Pool), which supports high-frequency tool invocation across parallel training workers.
Our contributions are summarized as follows:
\begin{itemize}[label=\raisebox{0.2ex}{\tiny$\bullet$}]
\item \textbf{Technical Contributions}. We introduce EDP that enhances exploration capability, enabling more effective policy refinement during RL.
We further propose MAR that dynamically balances heterogeneous quality metrics, effectively mitigating reward hacking in multi-objective optimization.
\item \textbf{Engineering Contribution}. We design a globally shared MC-Pool that supports distributed, high-frequency, and asynchronous GPU-based tool invocation, providing reliable infrastructure for training IR agents at scale.
\item \textbf{SOTA Performance}. TIR-Agent achieves over 2.5$\times$ inference speedup compared to training-free agents, and outperforms 12 baselines across both in-domain and out-of-domain settings.
\end{itemize}

\section{Related Work}

\subsection{Image Restoration}
As a fundamental research in computer vision, IR has been widely deployed in professional software and mobile ecosystems, such as Adobe Photoshop's Neural Filters~\cite{adobe_neural_filters} and Topaz Photo AI~\cite{topaz_photo_ai}.
The learning policy has evolved through three stages: \textit{restoring specific degradations}, \textit{unified models for unknown composite degradations}, \textit{agents to organize specialized models}.
Early IR methods adopt a task-specific paradigm, where models are trained for a predefined degradation.
Network architectures have evolved from CNN-based models~\cite{ir_cnn_sr, ir_cnn_res_dn, ir_cnn_res_sr} to GAN-based networks~\cite{ir_esrgan}, and Transformer-based architectures such as SwinIR~\cite{ir_swinir} and Uformer~\cite{ir_uformer}.
The models are supported by a unified framework BasicSR \cite{ir_basicsr}.
Later, motivated by the fact that most images often suffer from unknown and combined degradations, all-in-one restoration models are trained to handle multiple corruption types.
Mainstream approaches adopt dedicated strategies, such as contrastive learning~\cite{aio_airnet}, degradation classification~\cite{aio_dc} for feature differentiation, learnable prompts~\cite{aio_prompt}, adapters~\cite{aio_inject}, query injections~\cite{aio_multimodal}, and data scheduling~\cite{aio_foundir2}.
Besides, recent agent frameworks leverage VLM to assess degradation types~\cite{iragent_depictqa} and organize specialized tools~\cite{iragent_chain, iragent_hybrid}. 
These systems typically rely on planning mechanisms to determine the order of restoration subtasks, ranging from GPT-based selection \cite{iragent_depictqa}, prior knowledge-based staging \cite{jiang2025multi}, to greedy strategies \cite{iragent_greedy}.
However, current tool selection is inefficient due to exhaustive traversal~\cite{iragent_restoreagent} and often yields suboptimal performance when relying solely on LLM-based reasoning~\cite{iragent_greedy}.
To overcome these challenges, we investigate training the image restoration agent. 
A closely related work is SimpleCall \cite{simplecall}, which introduces a lightweight multilayer perceptron (MLP)-based IR agent. 
However, its MLP-based framework lacks context-aware reasoning, \textit{limiting its ability to determine whether and when to invoke super-resolution tools}.
With improved tool invocation and training strategies, TIR-Agent achieves superior efficiency and multi-metric performance.
%
%

\begin{figure}[tb!]
    \centering
    \includegraphics[width=1.0\linewidth]{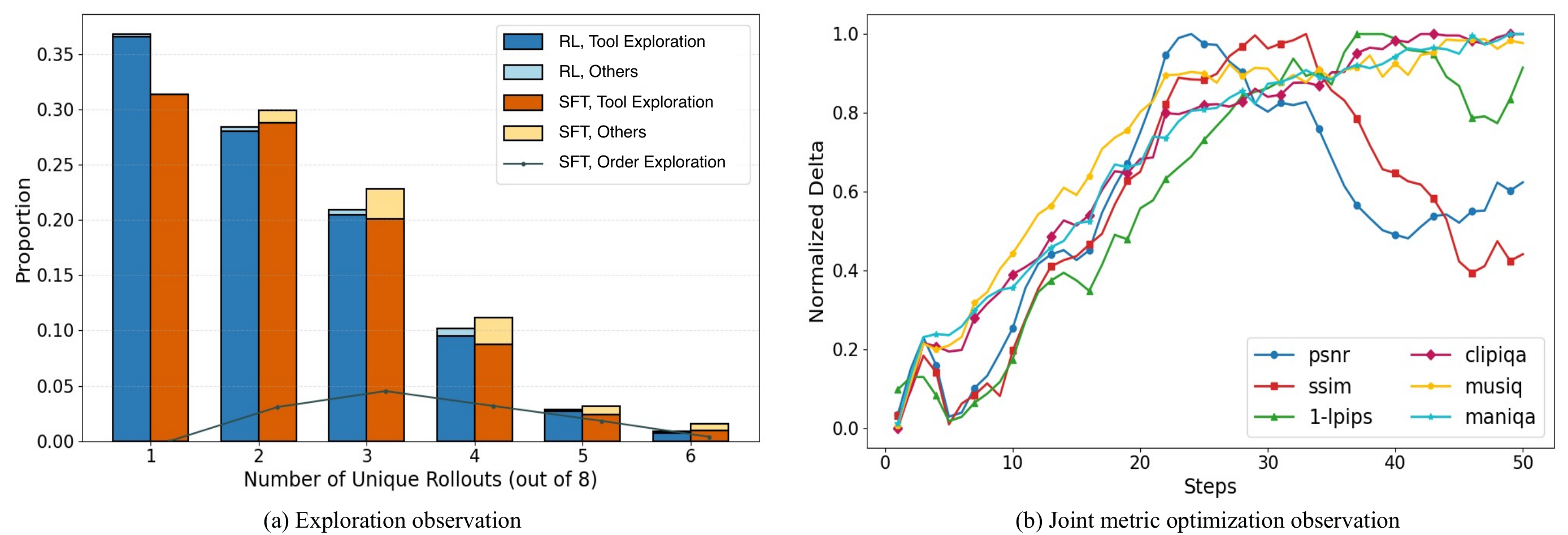}
    \caption{Observation of trajectory diversity, tool selection distribution, and optimization trend.
    (a) Statistics of the 8 rollouts per sample, including trajectory diversity (distinct trajectories), order diversity (same tasks with different execution orders, line), and model (tool) diversity (identical task sequences, different tool sequences).
    (b) Convergence trends of evaluation metrics during 50 RL training steps.
    }
    \label{fig:observation}
\end{figure}

\subsection{Vision-Language Agent}
Recently, increasing efforts have focused on developing vision-language agent frameworks for diverse visual tasks through tool composition.
Representative frameworks such as Visual Programming \cite{workflow_program} and ViperGPT \cite{workflow_vipergpt} translate queries into executable programs.
Similarly, Visual ChatGPT \cite{workflow_visualgpt} and HuggingGPT \cite{workflow_hugginggpt} utilize VLM-coordinated planning-execution pipelines to integrate external visual models and learned value functions \cite{brohan2023can} for grounded decision making.
%

%
To enhance multimodal reasoning capability, recent studies employ ``think with image'' paradigm.
These methods employ interleaved  chain-of-thought~\cite{imthink_deepeyes, imthink_deepeyes_v2}, curiosity-driven reward designs~\cite{su2025pixel}, long-horizon reasoning~\cite{imthink_videotraining}, and adaptive tool invocation~\cite{imthink_adatooler} for reasoning. 
Notably, recent studies have demonstrated that joint text-vision pre-training~\cite{imthink_kimi} significantly enhances multimodal reasoning capability.
Besides, a wider range of tools is introduced to encourage explicit exploration of visual evidence~\cite{imthink_octopus, imthink_videoqa}.
Considering that real images often suffer from diverse degradations~\cite{tang2025robust}, we investigate the agent framework.
%

\begin{figure}[t]
	\centering
	\includegraphics[width=1.0\textwidth]{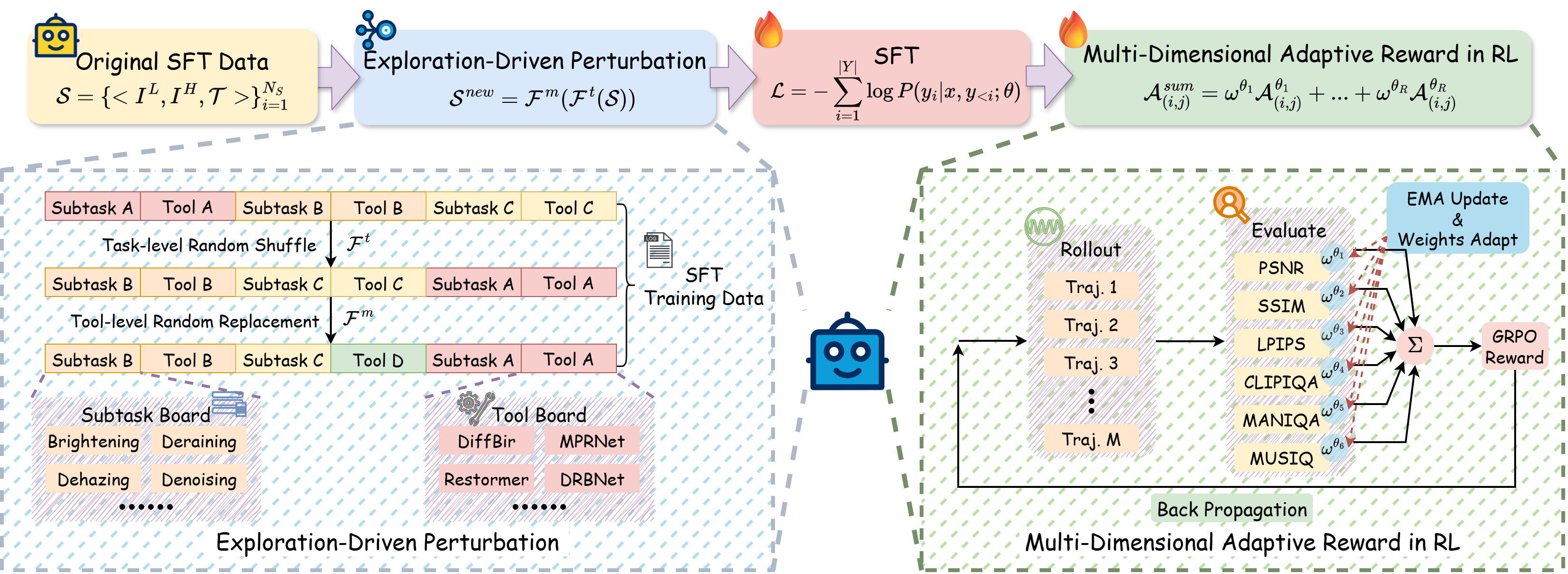}
	\caption{\textbf{Training pipeline of TIR-Agent.} The process begins with \textbf{SFT Data Generation}, followed by \textbf{Exploration-Driven Perturbation} to enhance data diversity via shuffled task sequences and tool substitutions. Following \textbf{SFT}, the agent undergoes \textbf{RL with a Multi-Dimensional Adaptive Reward} mechanism, which dynamically balances metric weights based on their EMA trends to optimize the restoration policy.}
	\label{pipeline}
\end{figure}

\section{Methodology}
\label{sec:method}

We present TIR-Agent, a trainable image restoration agent built upon
Qwen3-VL-8B-Instruct~\cite{bai2025qwen3vl}.
Given a degraded low-quality image $I^L$, TIR-Agent iteratively perceives the current image state, selects a restoration task (\textit{e.g.}, denoising, deblurring), and invokes a specialist tool to execute it.
As illustrated in Fig.~\ref{pipeline}, the agent is trained in two stages, where SFT learns a base tool-calling policy from demonstration trajectories, followed by
RL to refine this policy through
trial-and-error interaction with the FR and NR feedback.
We formulate IR as a finite-horizon decision process.
At each step $k$, the agent observes the image state sequence $\{s_0, s_1, \cdots, s_k\}$ ($s_0 = I^{L}$), and takes
an action $a_k = (\tau_k, m_k)$, where $\tau_k$ denotes the selected
restoration task and $m_k$ the specialist model.
Applying tool $m_k$ yields the next state $s_{k+1} = m_k(s_k)$.
The agent's policy $\pi_\theta(a_k \mid s_k, h_k)$ is parameterized by
the VLM, where $h_k = \{(a_j, s_j)\}_{j<k}$ is the interaction history.
The training objective is to learn $\pi_\theta$ that maximizes the
expected image quality of the final restored output.
Since the pretrained VLM lacks image quality perception and restoration tool selection capabilities, we construct a demonstration dataset
$\mathcal{S} = \{<I^{L}, I^{H}, \mathcal{T}>\}_{i=1}^{N_{S}}$.
Each trajectory $\mathcal{T}_i = \{(\tau_k, m_k)\}_{k=1}^{K_i}$ records
a sequence of task-tool decisions.
Trajectories are generated using training-free pipeline~\cite{zuo2025kagent}.

\subsection{Observations}
Before presenting our technical designs, we provide empirical observations from the vanilla SFT$+$RL setting that reveal two primary challenges in training an IR agent. These observations
directly motivate our proposed exploration-driven perturbation (EDP,
\S\ref{sec:edp}) and multi-dimensional adaptive reward (MAR,
\S\ref{sec:mar}).

\noindent\textbf{Observation 1: Limited exploration capability.}
As shown in Fig.~\ref{fig:observation}, during RL training, more than 35\% of the trajectories are identical, and a similar phenomenon is observed for SFT model.
%
%
Moreover, as shown in Fig.~\ref{fig:observation}(a), among different trajectories, only about 10\% differ in the task execution order, while trajectories with the same tasks but different tool selections are even fewer.
\noindent\textbf{Observation 2: Difficulty in joint multi-metric
optimization.}
Image quality assessment inherently involves multiple complementary
metrics (\textit{e.g.}, PSNR, SSIM, LPIPS).
When naively combining them as a scalar reward, we observe that the
agent tends to exploit partial metrics at the expense of others.
This occurs because heterogeneous metrics have different scales, sensitivities, and optimization landscapes, making uniform weighting unstable as training progresses.

\subsection{Exploration-Driven Perturbation}
\label{sec:edp}
To enhance exploration capability, we apply random perturbation to the SFT data to increase the diversity of task scheduling and tool invocation after fine-tuning.
The function for task scheduling, denoted as $\mathcal{F}^{t}$, is defined as:
\begin{equation}
\mathcal{F}^{t}(\mathcal{S}) = \mathcal{S} \cup \mathcal{G}(\mathcal{S}'), \; S' = \mathbbm{I}^{\alpha^t}(\mathcal{S}),
\end{equation}
where $\mathcal{S}' \subseteq \mathcal{S}$ is obtained by independently sampling elements in $\mathcal{S}$ with a probability $\alpha^t \in [0, 1]$.
For each sampled instance $\mathcal{S}_{i}' = <I^{L}_i, I^{H}_i, \mathcal{T}_i>$ in $\mathcal{S}'$, we apply perturbation operator $\mathcal{G}(\cdot)$ as:
\begin{equation}
\mathcal{G}(\mathcal{S}_{i}) = <I^{L}_i, I^{H}_i, \phi(\mathcal{T}_i)>,
\end{equation}
where $\phi(\cdot)$ denotes a random permutation over the order of the trajectory.
Next, we illustrate the tool-level perturbation.
For a given restoration task $t$, let $\mathcal{P}(m|t)$ denote the original tool selection distribution over candidate tools $m$.
To encourage exploration, we leverage function $\mathcal{F}^m$ to construct a perturbed distribution $\mathcal{P}^{m}(m|t)$ as:
\begin{equation}
\mathcal{P}^{m}(m|t) = (1-\alpha^m) \mathcal{P}(m|t) + \alpha^m U(m|t),
\end{equation}
where $U(m|t)$ denotes a uniform distribution over all valid tools for task $t$, and $\alpha^m \in [0, 1]$ is the perturbation probability.
This formulation interpolates between the original tool preference and uniform exploration, preventing the policy from collapsing to deterministic tool selection.
Finally, combining task-level and tool-level perturbations, the overall dataset $\mathcal{S}^{new}$ for SFT is defined as:
\begin{equation}
\mathcal{S}^{new} = \mathcal{F}^{m}(\mathcal{F}^{t}(\mathcal{S})).
\end{equation}

\subsection{Multi-Dimensional Adaptive Reward}
\label{sec:mar}
Image quality assessment inherently involves multiple metrics, which are often difficult to optimize jointly.
To address this issue, we propose a multi-dimensional adaptive reward mechanism that dynamically adjusts the optimization weight of each metric, encouraging the model to focus on the most under-optimized objective during training.
Given a batch of samples, taking metric $\theta_1$ as an example, we measure its relative performance change by comparing the current batch reward $r^{\theta_1}$ with its exponential moving average (EMA) $\overline{r^{\theta_1}}$.
Specifically, we define the clipped deviation score as:
\begin{equation}
    \hat{\omega}^{\theta_1} = 1 - \text{clip}\left(\frac{r^{\theta_1} - \overline{r^{\theta_1}}}{\overline{r^{\theta_1}}}, -\epsilon, \epsilon\right),
\end{equation}
where the clipping operation limits extreme fluctuations and stabilizes training.
When the current reward drops below its EMA, the weight increases, encouraging the policy to allocate more optimization effort to this metric.
The EMA is updated via:
\begin{equation}
    \overline{r^{\theta_1}} = (1 - \beta) r^{\theta_1} + \beta \overline{r^{\theta_1}},
\end{equation}
which provides a smoothed historical performance and reduces sensitivity to batch noise.
The final metric weights are normalized via the softmax function:
\begin{equation}
    \omega^{\theta_1} = \text{softmax}(\mathcal{W}), \;
 \mathcal{W} = \{\omega^{\theta_1}, ..., \omega^{\theta_R}\}.
\end{equation}
Next, inspired by~\cite{liu2026gdpo}, we compute the advantage for each metric in a decoupled manner before aggregation:
\begin{equation}
    \mathcal{A}_{(i, j)}^{\theta_1} = \frac{r_{(i, j)}^{\theta_1} - mean\{r^{\theta_1}_{(i, 1)}, ..., r^{\theta_1}_{(i, g)}\}}{std\{r^{\theta_1}_{(i, 1)}, ..., r^{\theta_1}_{(i, g)}\}},
\end{equation}
where $g$ denotes the size of a group of rollouts.
The decoupled advantages eliminate scale discrepancies, and the normalized advantages are subsequently combined via weighted summation as follows:
\begin{equation}
    \mathcal{A}_{(i, j)}^{sum} = \omega^{\theta_1} \mathcal{A}_{(i, j)}^{\theta_1} + ... + \omega^{\theta_R} \mathcal{A}_{(i, j)}^{\theta_R}.
\end{equation}
\subsection{Globally Shared MC-Pool}
Beyond algorithmic improvements, we develop a globally shared MC-Pool to support high-concurrency, high-frequency, GPU-based restoration model invocation during RL training.
During agentic RL training, the sampling is asynchronous, each step may trigger up to $b \times g$ concurrent tool invocations, where $b$ denotes batch size and $g$ is the number of a group of rollouts per sample.
Such large-scale concurrent execution leads to severe GPU memory cost and unpredictable runtime failures if unmanaged.
To address this issue, we develop a globally-shared pool to manage the resource set
$\mathcal{M} = \{M_1, M_2, ..., M_N\}$,
where each $M$ represents an independent MCP service instance bound to a specific GPU, which is deployed in a distributed manner.
Tool invocation requests are assigned to available instances through a locking-based allocation operator, which guarantees mutual exclusion during execution and releases the resource upon completion.
Further, we develop a retry mechanism (up to three attempts) to improve robustness against uncontrollable communication failures.
Overall, the MC-Pool provides a stable and scalable execution infrastructure that enables controlled high-concurrency scheduling and GPU-based restoration during RL training.

\begin{table}[!t]
\renewcommand{\arraystretch}{1.1}
\centering
\fontsize{6.0pt}{7.0pt}\selectfont
\setlength{\tabcolsep}{2pt}
\caption{Quantitative comparison of multiple-degradation IR tasks on three subsets (Group A, B, and C) from the MiO100 dataset against previous sota methods. The top three performances of each metric are marked in \textbf{bold}, \underline{underline}, \textit{italic}, respectively.}
\label{table:performance_open-source}
\resizebox{0.98\textwidth}{!}{
\begin{tabular}{>{\centering\arraybackslash}m{1.0cm}|>{\centering\arraybackslash}m{1.2cm}|m{2.0cm}|c c c c c c|c}
\toprule
\textbf{Dataset} & \textbf{Type} & \textbf{Method} &
\textbf{PSNR}$\uparrow$ & \textbf{SSIM}$\uparrow$ & \textbf{LPIPS}$\downarrow$ &
\textbf{MANIQA}$\uparrow$ & \textbf{CLIPIQA}$\uparrow$ & \textbf{MUSIQ}$\uparrow$ & \textbf{Time(s)}$\downarrow$ \\
\midrule
\midrule
\multirow{10}{*}{Group A}

& \multirow{6}{*}{All-in-One} & AirNet \cite{aio_airnet} & 19.13 & 0.6019 & 0.4283 & 0.2581 & 0.3930 & 42.46 & 1.75 \\  
&  & PromptIR \cite{potlapalli2023promptir} & 20.06 & 0.6088 & 0.4127 & 0.2633 & 0.4013 & 42.62 & 1.26 \\
&  & MiOIR \cite{kong2024towards} & 20.84 & 0.6558 & 0.3715 & 0.2451 & 0.3933 & 47.82 & 0.69 \\
&  & DA-CLIP \cite{luo2023controlling} & 19.58 & 0.6032 & 0.4266 & 0.2418 & 0.4139 & 42.51 & 44.87 \\  
&  & InstructIR \cite{conde2024instructir} & 18.03 & 0.5751 & 0.4429 & 0.2660 & 0.3528 & 45.77 & 1.45 \\ 
&  & AutoDIR \cite{jiang2024autodir} & 19.64 & 0.6286 & 0.3967 & 0.2500 & 0.3767 & 47.01 & 128.64 \\ 
\cline{2-10}
& \multirow{4}{*}{Agent} 
& AgenticIR \cite{iragent_depictqa} & \textit{21.04} & \underline{0.6818} & 0.3148 & 0.3071 & 0.4474 & 56.88 & 184.57 \\ 
&  & MAIR \cite{jiang2025multi} & 21.02 & 0.6715 & \underline{0.2963} & \textit{0.3330} & \textit{0.4751} & \textit{59.19} & - \\ 
&  & 4KAgent \cite{zuo2025kagent} & \underline{21.48} & \textit{0.6720} & \textit{0.3019} & \underline{0.3748} & \textbf{0.5544} & \underline{63.19} & 333.14 \\
&  & 
    \cellcolor{gray!30} TIR-Agent & \cellcolor{gray!30} \textbf{22.07} & \cellcolor{gray!30} \textbf{0.6935} & \cellcolor{gray!30} \textbf{0.2874} & \cellcolor{gray!30} \textbf{0.3907} & \cellcolor{gray!30} \underline{0.5454} & \cellcolor{gray!30} \textbf{63.69} & \textbf{61.58} \\
\midrule

\multirow{10}{*}{Group B}

& \multirow{6}{*}{All-in-One} & AirNet \cite{aio_airnet} & 19.31 & 0.6567 & 0.3670 & 0.2882 & 0.4274 & 47.88 & 1.83 \\  
&  & PromptIR \cite{potlapalli2023promptir} & 20.47 & 0.6704 & 0.3370 & 0.2893 & 0.4289 & 48.10 & 1.30 \\
&  & MiOIR \cite{kong2024towards} & 20.56 & 0.6905 & 0.3243 & 0.2638 & 0.4330 & 51.87 & 1.09 \\
&  & DA-CLIP \cite{luo2023controlling} & 18.56 & 0.5946 & 0.4405 & 0.2435 & 0.4154 & 43.70 & 47.41 \\  
&  & InstructIR \cite{conde2024instructir} & 18.34 & 0.6235 & 0.4072 & 0.3022 & 0.3790 & 50.94 & 1.50 \\ 
&  & AutoDIR \cite{jiang2024autodir} & 19.90 & 0.6643 & 0.3542 & 0.2534 & 0.3986 & 49.64 & 174.48 \\ 
\cline{2-10}
& \multirow{4}{*}{Agent} 
& AgenticIR \cite{iragent_depictqa} & 20.55 & \underline{0.7009} & 0.3072 & 0.3204 & 0.4648 & 57.57 & 201.25 \\ 
&  & MAIR \cite{jiang2025multi} & \textit{20.92} & \textit{0.7004} & \underline{0.2788} & \textit{0.3544} & \textit{0.5084} & \textit{60.98} & - \\ 
&  & 4KAgent \cite{zuo2025kagent} & \underline{20.95} & 0.6727 & \textit{0.3017} & \underline{0.3734} & \textbf{0.5505} & \underline{62.69} & 409.49 \\ 
&  & 
\cellcolor{gray!30} TIR-Agent & \cellcolor{gray!30} \textbf{22.80} & \cellcolor{gray!30} \textbf{0.7340} & \cellcolor{gray!30} \textbf{0.2725} & \cellcolor{gray!30} \textbf{0.3997} & \cellcolor{gray!30} \underline{0.5488} & \cellcolor{gray!30} \textbf{65.24} & \cellcolor{gray!30} \textbf{71.09} \\
\midrule

\multirow{10}{*}{Group C}

& \multirow{6}{*}{All-in-One} & AirNet \cite{aio_airnet} & 17.95 & 0.5145 & 0.5782 & 0.1854 & 0.3113 & 30.12 & 2.07 \\  
&  & PromptIR \cite{potlapalli2023promptir} & 18.51 & 0.5166 & 0.5756 & 0.1906 & 0.3104 & 29.71 & 1.36 \\
&  & MiOIR \cite{kong2024towards} & 15.63 & 0.4896 & 0.5376 & 0.1717 & 0.2891 & 37.95 & 2.06 \\
&  & DA-CLIP \cite{luo2023controlling} & 18.53 & 0.5320 & 0.5335 & 0.1916 & 0.3476 & 33.87 & 59.64 \\  
&  & InstructIR \cite{conde2024instructir} & 17.09 & 0.5135 & 0.5582 & 0.1732 & 0.2537 & 33.69 & 1.90 \\ 
&  & AutoDIR \cite{jiang2024autodir} & 18.61 & 0.5443 & 0.5019 & 0.2045 & 0.2939 & 37.86 & 240.34 \\ 
\cline{2-10}
& \multirow{4}{*}{Agent} 
& AgenticIR \cite{iragent_depictqa} & 18.82 & 0.5474 & 0.4493 & 0.2698 & 0.3948 & 48.68 & 249.70 \\ 
&  & MAIR \cite{jiang2025multi} & \textit{19.42} & \textit{0.5544} & \underline{0.4142} & \textit{0.2798} & \textit{0.4239} & \textit{51.36} & - \\ 
&  & 4KAgent \cite{zuo2025kagent} & \textbf{19.77} & \underline{0.5629} & \textit{0.4271} & \underline{0.3545} & \underline{0.5233} & \underline{55.56} & 555.12 \\
&  & \cellcolor{gray!30} TIR-Agent & \cellcolor{gray!30} \underline{19.53} & \cellcolor{gray!30} \textbf{0.5643} & \cellcolor{gray!30} \textbf{0.4120} &  \cellcolor{gray!30} \textbf{0.3739} & \cellcolor{gray!30} \textbf{0.5495} & \cellcolor{gray!30} \textbf{56.93} & \cellcolor{gray!30} \textbf{98.92} \\
\bottomrule
\end{tabular}}
\end{table}

\section{Experiments}
\label{sec:experiment}
\subsection{Implementation Details}
%
%
We adopt Qwen3-VL-8B-Instruct~\cite{bai2025qwen3vl} as the VLM and train it in two stages: SFT and RL.
In the SFT stage, the model is fully fine-tuned for 3 epochs on 8 NVIDIA A100 GPUs, with a learning rate of 1e-5.
The batch size is 1 with gradient accumulation of 4, and a cosine learning rate scheduler is applied.
In the RL stage, we employ 8 NVIDIA A100 GPUs to train the VLM, with a batch size of 64, 8 rollouts per sample, and a learning rate of 5e-6. 
The maximum number of parallel rollouts is capped at 128 to limit memory overhead.
In addition, we deploy 32 NVIDIA RTX 3090 GPUs as the MC-Pool to serve restoration model inference during rollout sampling.

\subsection{Datasets and Evaluation}
In the SFT stage, we construct 39,827 training samples containing trajectories generated by 4KAgent~\cite{zuo2025kagent}.
In the RL stage, we build 4,000 image pairs for training.
The images are sampled from ImageNet~\cite{deng2009imagenet} and are completely different from the test set to avoid data leakage. 
To obtain high-quality images, we filter 10k images based on several criteria, including image smoothness, aesthetic score~\cite{schuhmann2022improvedaestheticpredictor}, border presence, and overall image quality~\cite{you2024depicting}.
Following~\cite{iragent_depictqa}, we generate degraded images from these clean images. 
Notably, since the depth estimation used for adding haze is not publicly available, we replace it with Depth-Anything~\cite{yang2024depth}.
All evaluation experiments are conducted on MiO100~\cite{iragent_depictqa}.
Following previous works~\cite{iragent_depictqa, zuo2025kagent}, we employ three FR metrics (PSNR, SSIM, LPIPS) and three NR metrics (MANIQA, CLIPIQA, MUSIQ) to evaluate restoration performance.

\begin{table*}[!tb]
\renewcommand{\arraystretch}{1.1}
\centering
\fontsize{6.0pt}{7.0pt}\selectfont
\setlength{\tabcolsep}{2pt}
\caption{Quantitative comparison of multiple-degradation IR tasks on three subsets (Group A, B, and C) from the MiO100 dataset against proprietary models.}
\label{table:performance_closed-source}

\resizebox{1.0\textwidth}{!}{
\begin{tabular}{>{\centering\arraybackslash}m{1.0cm}|m{2.17cm}|c c c c c c|c}
\toprule
\textbf{Dataset} & \textbf{Method} &
\textbf{PSNR}$\uparrow$ & \textbf{SSIM}$\uparrow$ & \textbf{LPIPS}$\downarrow$ &
\textbf{MANIQA}$\uparrow$ & \textbf{CLIPIQA}$\uparrow$ & \textbf{MUSIQ}$\uparrow$ & \textbf{Time(s)}$\downarrow$ \\
\midrule
\midrule
\multirow{4}{*}{Group A}

& 
GPT-5.2 \cite{gpt52} &21.33 &0.6620 &0.3639 & \underline{0.3385}&\underline{0.4870} &\textit{57.45} & 101.32\\
& Claude Opus 4.5 \cite{claude_opus45}& \textit{21.47} & \underline{0.6999}& \textit{0.3264} & 0.3073& 0.4538 & 55.89&97.71\\ 
& Gemini 3.0 Pro \cite{gemini3provision2025}&\textbf{22.23} &\textbf{0.7085} & \underline{0.3151} &\textit{0.3149} & \textit{0.4573}&\underline{57.95} &99.59 \\
& \cellcolor{gray!30} TIR-Agent & \cellcolor{gray!30} \underline{22.07} & \cellcolor{gray!30} \textit{0.6935} & \cellcolor{gray!30} \textbf{0.2874} & \cellcolor{gray!30} \textbf{0.3907} & \cellcolor{gray!30} \textbf{0.5454} & \cellcolor{gray!30} \textbf{63.69} & \cellcolor{gray!30} \textbf{61.58} \\
\midrule

\multirow{4}{*}{Group B}

& 
GPT-5.2 \cite{gpt52} & \textit{20.84}& 0.6627&0.3557 & \underline{0.3369}& 0.4863& \underline{58.27}& 110.65\\
& Claude Opus 4.5 \cite{claude_opus45}&20.46 &\underline{0.7051} &\underline{0.3145} & 0.3189& 0.4643& 56.24&112.52\\ 
& Gemini 3.0 Pro \cite{gemini3provision2025} &\underline{21.03} & \textit{0.6927}& \textit{0.3238}& \textit{0.3225}& 0.4764& \textit{57.82}&109.83 \\
& \cellcolor{gray!30} TIR-Agent & \cellcolor{gray!30} \textbf{22.80} & \cellcolor{gray!30} \textbf{0.7340} & \cellcolor{gray!30} \textbf{0.2725} & \cellcolor{gray!30} \textbf{0.3997} & \cellcolor{gray!30} \textbf{0.5488} & \cellcolor{gray!30} \textbf{65.24} & \cellcolor{gray!30} \textbf{71.09} \\
\midrule

\multirow{4}{*}{Group C}

& 
GPT-5.2\cite{gpt52} &\textit{19.53} &\textit{0.5759} &0.4441 &\underline{0.3218} & \underline{0.4604}&\underline{53.09} & 133.44\\
& Claude Opus 4.5\cite{claude_opus45} &\underline{20.27} &\underline{0.6043} &\textit{0.4228} &\textit{0.2660} &\textit{0.4132} & 48.12&137.40\\ 
& Gemini 3.0 Pro\cite{gemini3provision2025} &\textbf{20.38} &\textbf{0.6118} & \underline{0.4150}& 0.2577& 0.3973& \textit{48.62}& 148.51\\
& \cellcolor{gray!30} TIR-Agent & \cellcolor{gray!30} \textit{19.53} & \cellcolor{gray!30} 0.5643 & \cellcolor{gray!30} \textbf{0.4120} & \cellcolor{gray!30} \textbf{0.3739} & \cellcolor{gray!30} \textbf{0.5495} & \cellcolor{gray!30} \textbf{56.93} & \cellcolor{gray!30} \textbf{98.92} \\
\bottomrule
\end{tabular}
}
\end{table*}

\subsection{Comparison with SOTA Methods}

\noindent\textbf{Baselines}.
We compare our TIR-Agent against both all-in-one and agentic methods.
All-in-one methods include AirNet \cite{aio_airnet}, PromptIR \cite{potlapalli2023promptir}, MiOIR \cite{kong2024towards}, DA-CLIP \cite{luo2023controlling}, InstructIR \cite{conde2024instructir}, and AutoDIR \cite{jiang2024autodir}.
Agentic methods contain three training-free works, \textit{i.e.}, AgenticIR \cite{iragent_depictqa}, MAIR \cite{jiang2025multi}, 4KAgent \cite{zuo2025kagent}, and another trainable agentic method SimpleCall~\cite{simplecall}.
Notably, we further compare our method with proprietary SOTA models, including GPT-5.2\cite{gpt52}, Claude Opus 4.5\cite{claude_opus45}, and Gemini 3.0 Pro\cite{gemini3provision2025}.
%
%

\noindent\textbf{Main comparison}.
Tab.~\ref{table:performance_open-source} presents the comparison with 9 state-of-the-art methods under three settings on the MiO100~\cite{iragent_depictqa}. We have three observations.
1) \textit{\textbf{Effectiveness}}: Our TIR-Agent exhibits clear superiority, \textbf{achieving consistent improvements on both full-reference (FR) fidelity metrics and no-reference (NR) image quality metrics}.
2) \textit{\textbf{Efficiency}}: Compared with training-free agentic methods, TIR-Agent trains the VLM to directly select the optimal restoration model, leading to faster inference.
In particular, TIR-Agent \textbf{achieves at least 2.5$\times$ and 5$\times$ speedup over AgenticIR and 4KAgent}, respectively.
3) \textit{\textbf{IR Expertise}}: Compared with leading proprietary models, TIR-Agent outperforms them on most metrics, with \textbf{especially notable advantages on NR image quality metrics}.

\begin{figure}[!t]
    \centering
    \includegraphics[width=0.95\linewidth]{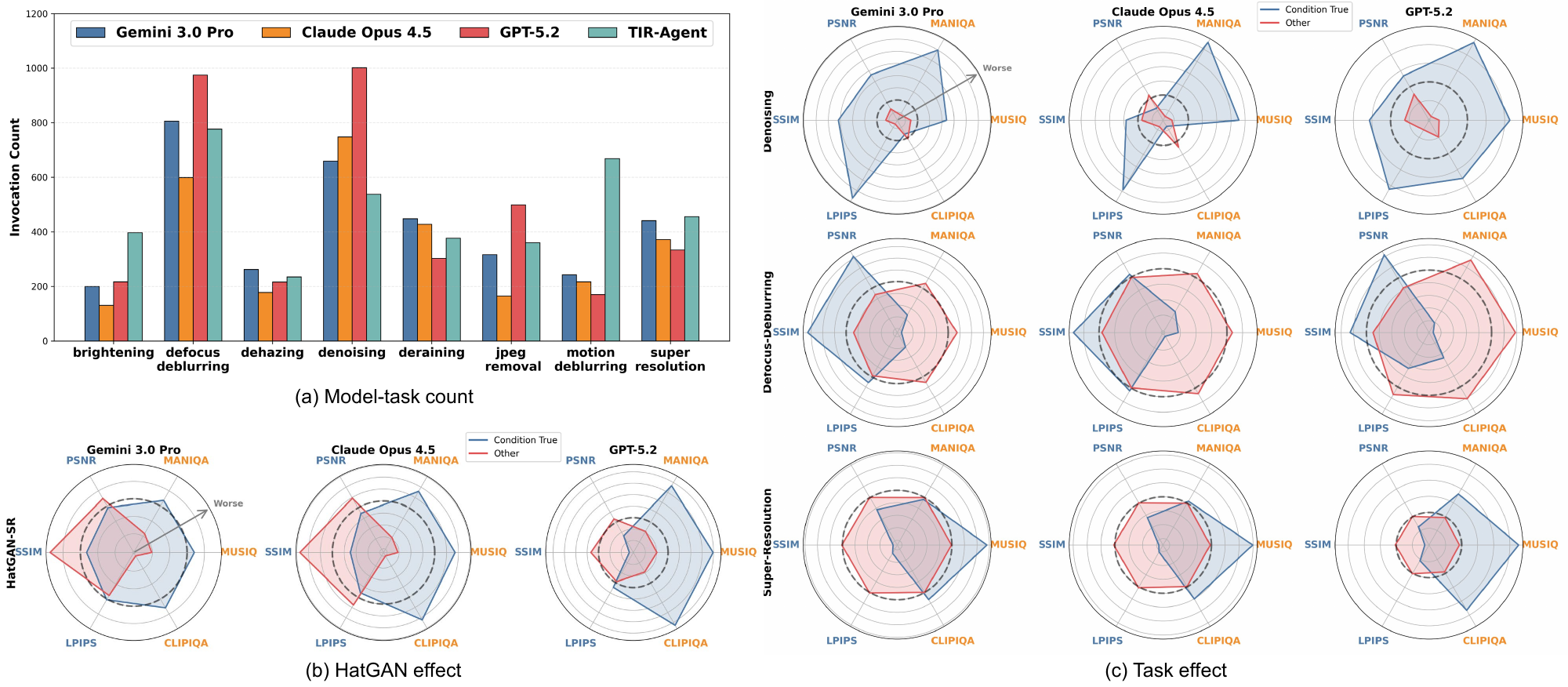}
    \caption{Analysis of proprietary models. (a) presents the invocation count of each IR tasks on MiO100. (b) presents the effect of HatGAN. (c) presents the effect of IR tasks. Note that for denoising, defocus deblurring, and HatGAN, the condition is proprietary model invokes more times of such task / model, super-resolution is the opposite. Then, we standardize the differences between these models and TIR-Agent in each metric.}
    \label{fig:task_tool_analyze}
\end{figure}

Next, we compare TIR-Agent with another trainable agentic method, SimpleCall~\cite{simplecall}, in Tab.~\ref{tab:simplecall}. 
Since SimpleCall is not publicly available and the degradation combinations differ, we report results under the partial settings described in the paper. 
We observe:
1) \textit{\textbf{Balance}}: SimpleCall$^{rr}$ drops significantly on FR metrics.
This phenomenon is consistent with our motivation for MAR, which aims to address the challenge of jointly optimizing heterogeneous image quality metrics.
\begin{wraptable}[8]{l}{0.57\linewidth}
    \vspace{-30pt}
    \centering
    \renewcommand{\arraystretch}{1.1}
    \caption{
        Comparing TIR-Agent with SimpleCall on five degradation combinations in MiO100's GroupA.
        Only these five combinations have complete results of $SimpleCall^{fr}$ and $SimpleCall^{rr}$.
    }
    \resizebox{0.57\textwidth}{!}{
    \begin{tabular}{c | cccccc | c}
    \toprule
    Method & \textbf{PSNR}$\uparrow$ & \textbf{SSIM}$\uparrow$ & \textbf{LPIPS}$\downarrow$ & \textbf{MANIQA}$\uparrow$ & \textbf{CLIPIQA}$\uparrow$ & \textbf{MUSIQ}$\uparrow$ & \textbf{Time(s)}$\downarrow$  \\
    \midrule \midrule
    SimpleCall$^{rr}$ & 19.51 & 0.6474 & 0.3650 & 0.3492 & 0.5252 & 63.00 & 80.20 \\
    SimpleCall$^{fr}$ & 22.51 & 0.7212 & 0.2928 & 0.3346 & 0.4448 & 59.65 & - \\
    TIR-Agent & \textbf{23.01} & \textbf{0.7421} & \textbf{0.2800} & \underline{0.3448} & \underline{0.4874} & \underline{61.85} & \textbf{54.38} \\
    \bottomrule
    \end{tabular}}
    \label{tab:simplecall}
\end{wraptable}
2) \textit{\textbf{Superiority}}: Compared with SimpleCall$^{fr}$, TIR-Agent achieves better performance on all metrics.
3) \textit{\textbf{Task scalability}}: SimpleCall trains an MLP to select restoration tasks and tools without contextual awareness, which prevents it from scheduling super-resolution tasks.
In contrast, TIR-Agent can naturally extend to arbitrary restoration tasks.
4) \textit{\textbf{Efficiency}}: Both methods directly select restoration models, but TIR-Agent achieves higher efficiency, indicating that it can obtain performance with fewer rounds of tool invocations.
%

\begin{table}[!tb]
\renewcommand{\arraystretch}{1.1}
\centering
\fontsize{6.0pt}{7.0pt}\selectfont
\setlength{\tabcolsep}{2pt}
  \caption{Quantitative comparison on out-of-domain degradation combinations across two categories: $D=2$ (mixtures of two degradations) and $D \ge 3$ (mixtures of three, four, five degradations). Details of degradations are provided in the supplementary material. The top three performances of each metric are marked in \textbf{bold}, \underline{underline}, \textit{italic} respectively.}
\resizebox{0.97\textwidth}{!}{
\begin{tabular}{>{\centering\arraybackslash}m{1.0cm}|>{\centering\arraybackslash}m{1.4cm}|m{2.0cm}|c c c c c c}

    \toprule
\textbf{Group} & 
\textbf{Type} & 
\textbf{Method} & 
\textbf{PSNR}$\uparrow$ & \textbf{SSIM}$\uparrow$ & \textbf{LPIPS}$\downarrow$ &
\textbf{MANIQA}$\uparrow$ & \textbf{CLIPIQA}$\uparrow$ & \textbf{MUSIQ}$\uparrow$ \\
    \midrule
    \midrule
    \multirow{10}[2]{*}{$D = 2$} & \multirow{7}{*}{All-in-One} & AirNet\cite{aio_airnet} & 19.05  & 0.5892  & 0.4454  & 0.2384  & 0.3030  & 37.59  \\
          & & PromptIR\cite{potlapalli2023promptir}  & 19.53  & 0.5947  & 0.4441  & 0.2445  & 0.3223  & 37.06  \\
          
          & & MiOIR(R) \cite{kong2024towards}& 18.93  & 0.6098  & 0.4131  & 0.2268  & 0.3336  & 39.73  \\
          & & MiOIR(U)\cite{kong2024towards} & 18.62  & 0.6092  & 0.4129  & 0.2309  & 0.3422  & 40.56  \\
          & & DA-CLIP\cite{luo2023controlling} & \textit{20.20}  & 0.6266  & 0.4062  & 0.2436  & 0.3371  & 40.19  \\
          & & InstructIR\cite{conde2024instructir} & 18.63  & 0.5622  & 0.4780  & 0.2622  & 0.3136  & 42.38  \\
          & & AutoDIR \cite{jiang2024autodir}& 19.18  & 0.6013  & 0.4046  & \underline{0.2980}  & 0.3646  & 50.31  \\
          \cline{2-9}
          & \multirow{3}{*}{Agent} & AgenticIR  \cite{iragent_depictqa}& \underline{21.54}  & \underline{0.7006}  & \underline{0.3492}  & 0.2762  & \textit{0.3833} & \textit{52.57} \\
          & & 4KAgent\cite{zuo2025kagent} & 20.17 & \textit{0.6488} & \textit{0.3519} & \textit{0.2936} & \textbf{0.4341} & \underline{55.65}  \\
          & & \cellcolor{gray!30} TIR-Agent & \cellcolor{gray!30} \textbf{23.19} & \cellcolor{gray!30} \textbf{0.7526} & \cellcolor{gray!30} \textbf{0.3316} & \cellcolor{gray!30} \textbf{0.3103} & \cellcolor{gray!30} \underline{0.4097}  & \cellcolor{gray!30} \textbf{57.63} \\
    \midrule
    \multirow{10}[2]{*}{$D \ge 3$} & \multirow{7}{*}{All-in-One} & AirNet\cite{aio_airnet} & 17.36  & 0.4387  & 0.6162  & 0.1477  & 0.2198  & 25.12  \\
          & & PromptIR\cite{potlapalli2023promptir}  & 17.86  & 0.4502  & 0.6071  & 0.1553  & 0.2356  & 25.33  \\
          & & MiOIR(R)\cite{kong2024towards} & 18.11  & 0.4719  & 0.6087  & 0.1540  & 0.2573  & 26.76  \\
          & & MiOIR(U)\cite{kong2024towards} & 18.10  & 0.4700  & 0.6067  & 0.1564  & 0.2817  & 26.98  \\
          & & DA-CLIP\cite{luo2023controlling} & 17.84  & 0.4536  & 0.6136  & 0.1729  & \textit{0.2843} & 28.06  \\
          & & InstructIR \cite{conde2024instructir} & 18.04  & 0.4664  & 0.6065  & 0.1509  & 0.2218  & 26.05  \\
          & & AutoDIR \cite{jiang2024autodir}& 18.12  & 0.4651  & 0.5668  & 0.1726  & 0.2429  & 30.13  \\
          \cline{2-9}
          & \multirow{3}{*}{Agent} & AgenticIR \cite{iragent_depictqa} & \textit{18.13} & \underline{0.5237}  & \textit{0.5429} & \textit{0.1777} & 0.2801  & \underline{38.00}  \\
          & & 4KAgent\cite{zuo2025kagent} & \underline{18.17}  & \textit{0.5072} & \underline{0.5384}  & \underline{0.1909}  & \underline{0.2989}  & \textit{37.58} \\
          & & \cellcolor{gray!30} TIR-Agent & \cellcolor{gray!30} \textbf{18.69} & \cellcolor{gray!30} \textbf{0.5287} & \cellcolor{gray!30} \textbf{0.5233} & \cellcolor{gray!30} \textbf{0.2079} & \cellcolor{gray!30} \textbf{0.3268} & \cellcolor{gray!30} \textbf{45.55} \\
    \bottomrule
    \end{tabular}%
  \label{tab:addlabel}%
  }
\end{table}%

\noindent\textbf{Comparison with proprietary models}.
In Tab.~\ref{table:performance_closed-source}, we compare TIR-Agent with three proprietary models.
We can observe that these models achieve competitive performance on FR metrics, but perform notably worse on NR metrics.
To better understand the underlying factors behind this phenomenon, we conduct a detailed analysis as shown in Fig.~\ref{fig:task_tool_analyze}.
1) \textit{\textbf{Task bias.}} As shown in Fig.~\ref{fig:task_tool_analyze} (a), these models prefer to select denoising and defocus deblurring, while invoke fewer motion deblurring and brightening.
It is reasonable, since real-world low-quality images usually lack professional annotations and are often directly attributed to noise.
2) \textit{\textbf{Task effect.}} Fig.~\ref{fig:task_tool_analyze} (c) analyzes the effect of denoising, defocus deblurring, and super-resolution. 
Excessive denoising degrades MANIQA, LPIPS, and MUSIQ, more defocus deblurring benefits NR metrics but harms PSNR/SSIM. 
Additionally, the lack of super-resolution invocation improves FR metrics but noticeably reduces NR metrics, especially MUSIQ.
Empirically, the low-resolution images may reduce the influence of high-frequency differences. 
3) \textit{\textbf{Tool effect.}} Without IR-specific training, proprietary models lack knowledge of tool differences for the same task.
For example, they prefer HatGAN for super-resolution, which improves FR metrics but degrades NR metrics as demonstrated in Fig.~\ref{fig:task_tool_analyze} (b).
More analysis about task and model is provided in the supplementary material.

\noindent\textbf{Out-of-domain degradation combination validation}.
%
%
To evaluate generalization capability, we conduct experiments on \textit{unseen degradation combinations} during training, with results reported in Tab.~\ref{tab:addlabel}. 
We observe:
1) \textit{\textbf{Superiority}}: Under the settings of $D = 2$ and $D \ge 3$, TIR-Agent achieves the best performance on 11 out of 12 results compared with nine baselines.
2) \textit{\textbf{Generalization capability}}: As the number of degradations increases, TIR-Agent shows more pronounced advantages on most metrics, particularly NR metrics, demonstrating its strong generalization capability.

\begin{figure}[!tb]
    \centering
    \includegraphics[width=0.98\linewidth]{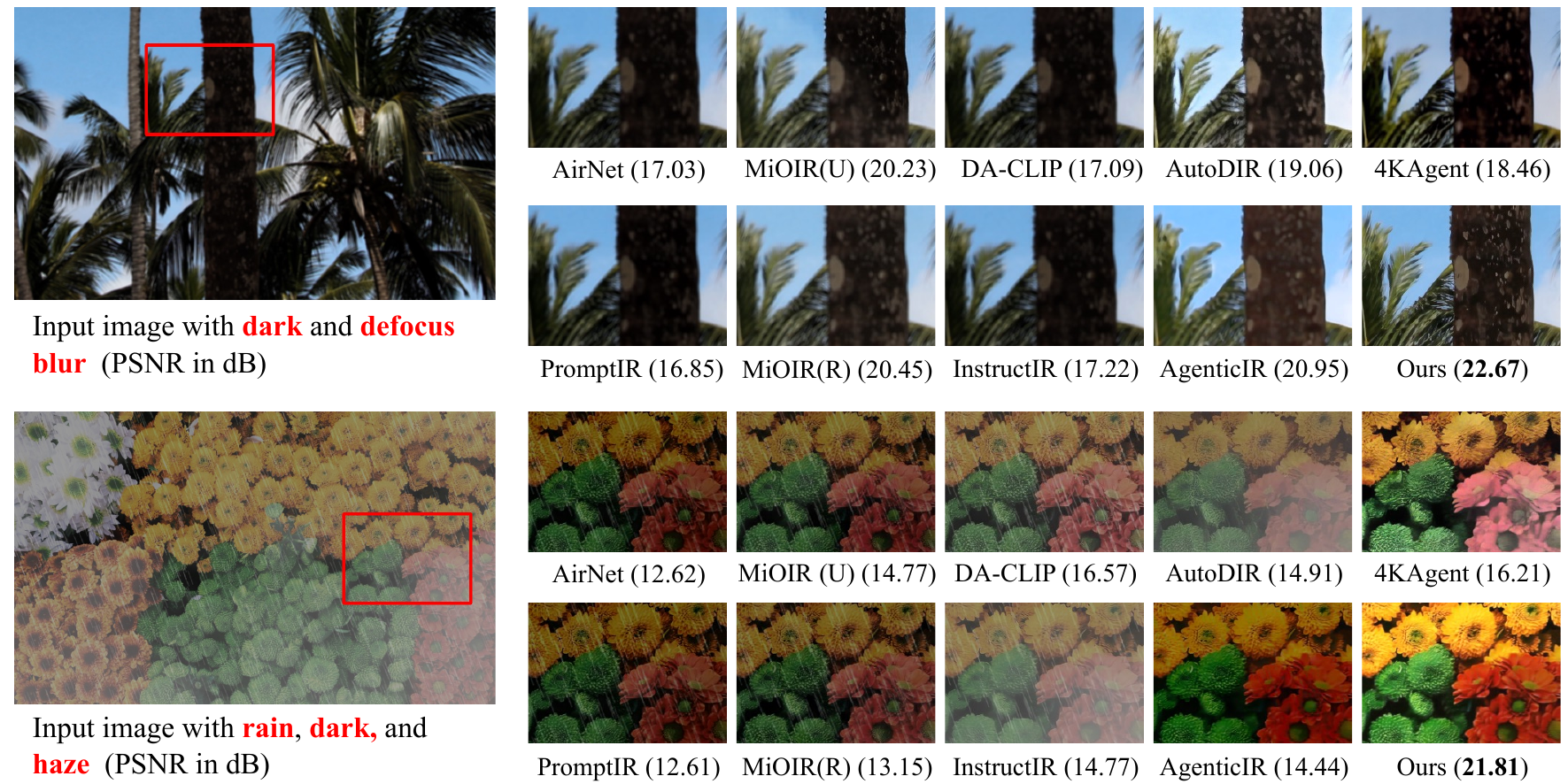}
    \caption{Visual comparison examples on MiO100 dataset.}
    \label{fig:comparison}
\end{figure}

\noindent\textbf{Qualitative comparison}.
According to the qualitative examples in Fig.~\ref{fig:comparison}, we observe that our method effectively addresses all degradations in the image and generates visually pleasing results.
Besides, comparing all-in-one models and agentic methods, we find that all-in-one models tend to perform well on dehazing, but struggle to handle dark, revealing a notable task bias.

\subsection{Ablation Study}
\noindent\textbf{Overall ablation}.
The overall ablation results of the training algorithm and optimizations are provided in Tab.~\ref{tab:ablation}.
We have the following three observations:
1) \textit{\textbf{Importance of SFT and RL}}.
Both SFT and RL are crucial.
SFT enables the VLM with the ability to understand image degradations, select appropriate restoration tasks and models, and follow the required output format.
\begin{wraptable}[12]{l}{0.57\columnwidth}
    \vspace{-30pt}
    \centering
    \caption{Ablation results of training algorithm and optimization strategy in TIR-Agent.}
    \renewcommand{\arraystretch}{1.1}
    \resizebox{0.57\textwidth}{!}{
    \begin{tabular}{c | cccccc}
    \toprule
    Method & \textbf{PSNR}$\uparrow$ & \textbf{SSIM}$\uparrow$ & \textbf{LPIPS}$\downarrow$ & \textbf{MANIQA}$\uparrow$ & \textbf{CLIPIQA}$\uparrow$ & \textbf{MUSIQ}$\uparrow$  \\
    \midrule \midrule
    TIR-Agent & 22.07 & 0.6935 & 0.2874 &  0.3907 & 0.5454 & 63.69 \\
    \midrule
    Vanilla & 20.71 & 0.6893 & 0.3501 & 0.3550 & 0.4481 & 56.62 \\
    w/o SFT & 19.49 & 0.6097 & 0.3288 & 0.3706 & 0.5239 & 59.51 \\
    w/o RL & 21.78 & 0.6812 & 0.3060 & 0.3319 & 0.5258 & 60.17 \\
    \midrule
    w/o EDP & 19.97 & 0.6227 & 0.2989 & 0.3756 & 0.5518 & 62.54 \\
    w/o $\alpha^{t}$ & 20.61 & 0.6501 & 0.2847 & 0.3745 & 0.5497 & 62.49 \\
    w/o $\alpha^{m}$ & 21.95 & 0.6749 & 0.3256 & 0.3860 & 0.5125 & 60.94 \\
    \midrule
    w/o MAR & 20.85 & 0.6792 & 0.2830 & 0.3797 & 0.5437 & 63.23 \\
    w/o decouple & 21.86 & 0.6776 & 0.3098 & 0.3660 & 0.5335 & 61.17 \\
    w/o $\omega^{\theta}$ & 20.51 & 0.6573 & 0.2995 & 0.3808 & 0.5462 & 63.15 \\
    \bottomrule
    \end{tabular}}
    \label{tab:ablation}
\end{wraptable}
Removing SFT severely degrades performance, especially on PSNR and SSIM.
Further, RL encourages the model to jointly improve all metrics on top of the SFT initialization.
2) \textit{\textbf{Effect of EDP}}.
Without EDP, training diversity quickly collapses, aggravating the imbalance among metrics.
In later stages of training, this even leads to drops in PSNR and SSIM.
3) \textit{\textbf{Effect of MAR}}.
Without MAR, the model achieves NR metrics close to TIR-Agent but still lags behind on FR metrics.

\begin{figure}[!tb]
    \centering
    \includegraphics[width=1.0\linewidth]{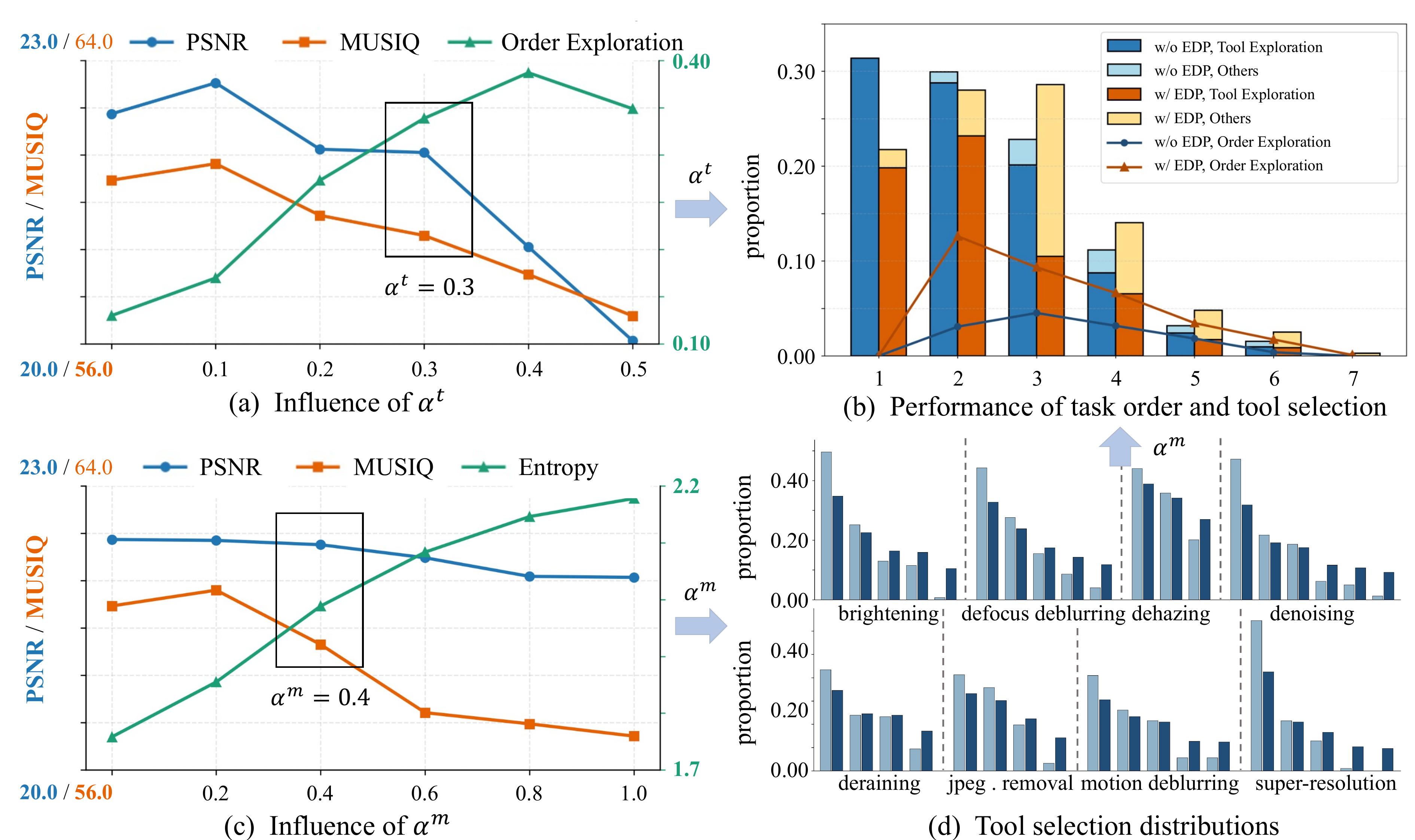}
    \caption{Influence of EDP and MAR. (a) Influence of $\alpha^t$ on metrics and order exploration (proportion of samples with same tasks with different execution orders). 
    (b) Performance of order and tool exploration. 
    (c) Influence of $\alpha^m$ for metrics and tool selection (average entropy of tool distribution for given task.) 
    (d) Tool selection distributions.}
    \label{fig:ab_sft}
\end{figure}

\noindent\textbf{Analysis of EDP}.
%
%
We provide a detailed analysis of the effect of EDP on the order schedule and model (tool) selection in Fig.~\ref{fig:ab_sft},
1) \textbf{fine-tuning with the EDP-adjusted SFT data significantly increases the proportion of trajectories that explore order scheduling}.
This trend is consistently observed in samples with 2–6 distinct rollouts among the 8 trajectories generated per sample.
Moreover, 2) \textbf{the bar plots indicate a substantial increase in model (tool) diversity across different rollouts}.
Specifically, Fig.~\ref{fig:ab_sft} (d) compares the tool selection distributions before and after applying EDP.
After EDP adjustment, the predicted distribution becomes smoother, and the usage probability of previously under-invoked tools is significantly increased.
\begin{figure}[!tb]
    \centering
    \includegraphics[width=0.98\linewidth]{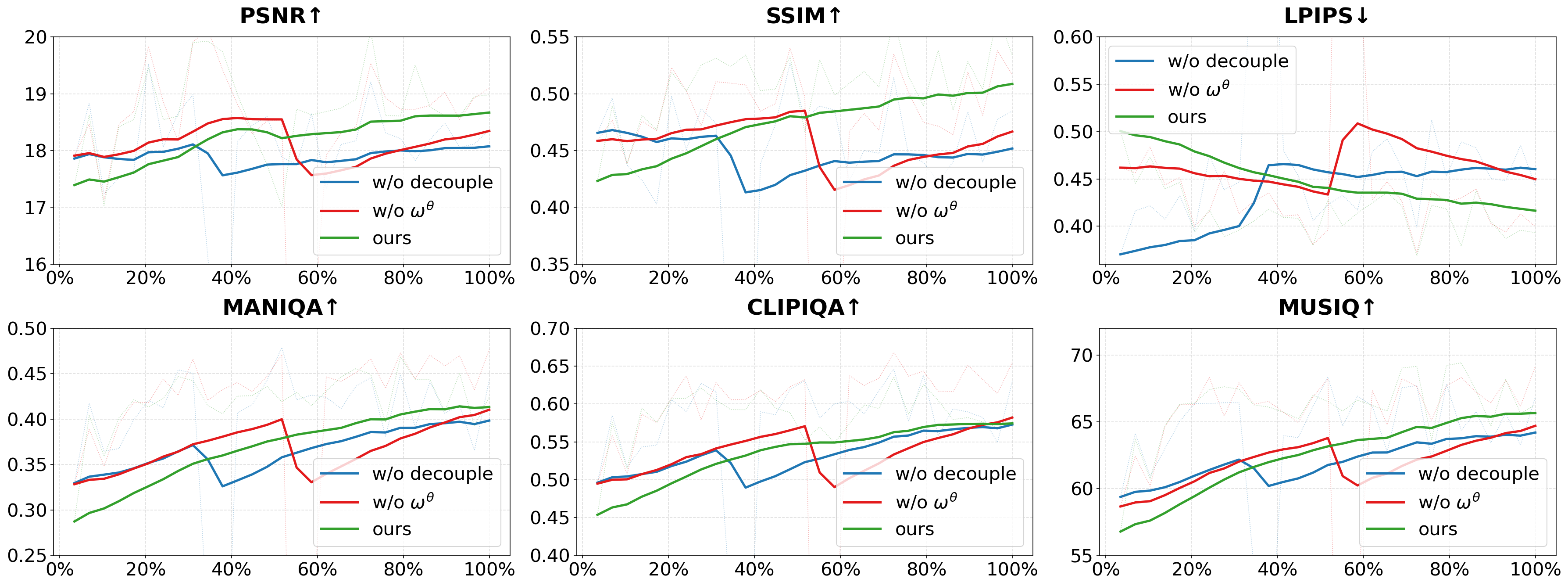}
    \caption{Training curves under different reward settings: w/o decouple, w/o $\omega ^ {\theta}$ and ours.}
    \label{fig:rl_curve}
\end{figure}

\noindent\textbf{Analysis of MAR}.
As shown in Fig.~\ref{fig:rl_curve}, using both $\omega ^ {\theta}$ and the decoupling strategy \textbf{enables the 6 metrics to converge in a coordinated manner during training, achieving the best performance on five of the metrics.}
Without decoupling, the model shows limited improvement on the 3 FR metrics. This is likely because, without normalization, the NR metrics have larger numerical scales and dominate the reward signal, thereby limiting the contribution of the FR metrics.
When $\omega ^ {\theta}$ is removed, the model performs relatively well on the three NR metrics, but a noticeable performance drop occurs midway through training. Although the metrics eventually recover, the FR metrics only return to their pre-drop levels, due to the lack of dynamic weight adaptation, which makes it difficult for the model to simultaneously optimize multiple equally weighted objectives.
Overall, these results highlight the importance of both decoupling and dynamic weighting in achieving stable and balanced multi-metric optimization.

\subsection{Hyperparameter Analysis}
%
%


%
We analyze the effect and sensitivity of two hyperparameters in TIR-Agent, $\alpha^{t}$ and $\alpha^{m}$.
As shown in Fig.~\ref{fig:ab_sft}(a), increasing $\alpha^{t}$ continuously improves the order diversity, with a slight drop at 0.5.
However, performance begins to degrade significantly when $\alpha^{t} >$ 0.3, especially for PSNR.
We attribute this to the fact that after shuffling the order, the model may encounter examples where low-quality images appear before high-quality ones, which makes the model confusing.
Therefore, we set $\alpha^{t}=$ 0.3 to balance order diversity and performance.
As shown in Fig.~\ref{fig:ab_sft} (c), increasing $\alpha^{m}$ significantly improves tool diversity (larger entropy, \textit{i.e.}, smoother distribution). 
Meanwhile, $\alpha^{m}$ has minimal impact on PSNR, indicating that different tools for the same restoration task mainly affect perceptual quality rather than fidelity. We set $\alpha^{m} =$ 0.4 to balance tool diversity and performance.
%


\section{Conclusion}
\label{sec:conclusion}
In this paper, we introduce TIR-Agent, a trainable vision-language agent that addresses the fundamental limitations of existing training-free IR agent frameworks.
Rather than relying on heuristic planning and exhaustive tool traversal, TIR-Agent learns a direct tool-calling policy that dynamically selects restoration tasks and models based on the current image state, through a two-stage training pipeline of SFT and RL.
To enable effective policy learning, we identify two key challenges in agentic RL for IR: limited exploration in restoration trajectories and unstable optimization across heterogeneous image quality metrics.
To address these issues, we introduce an EDP strategy that expands the diversity of task ordering and tool selection during SFT, and an MAR mechanism that dynamically balances multiple image quality metrics to prevent reward hacking and stabilize multi-objective optimization.
Extensive experiments demonstrate that TIR-Agent consistently outperforms existing all-in-one models and agentic frameworks on both in-domain and out-of-domain composite degradations.
Meanwhile, by directly selecting the restoration tool, TIR-Agent achieves over 2.5$\times$ faster inference speed.
These results highlight the effectiveness of learning-based tool orchestration for IR, and suggest that agentic RL provides a promising direction for building scalable and adaptive IR systems.

\section*{Acknowledgements}
This work is supported by the National Science and Technology Major Project (2023ZD0121403).

\clearpage



%
%
\bibliographystyle{splncs04}
\bibliography{main}
\end{document}